\documentclass{CVM}

\CVMsetup{
type      = {Research/Review Article},
doi       = {CVM.XXXX},
title     = {CORGI: \textbf{C}onsistency-Aware 3D D\textbf{o}g \textbf{R}econstruction from a Sin\textbf{g}le \textbf{I}mage in the Wild},
author    = {Yuxiao Wu$^{1,\dagger}$, Weile Li$^{1,\dagger}$, Boyi Zhu$^{1}$, Yumeng Liu$^{1}$\cor{}, Youcheng Cai$^{1}$\cor{},  XiaoMing Fu$^{1}$, and Ligang Liu$^{1}$},
runauthor = {Y. Wu, W. Li, B. Zhu, Y. Liu, Y. Cai, X. Fu, L. Liu},
abstract  = {
Reconstructing high-fidelity 3D models of highly articulated animals, such as dogs, from a single in-the-wild image remains a formidable challenge. In this paper, we introduce CORGI, a novel framework for consistency-aware 3D dog reconstruction from a single input image that completely eliminates the need for 3D supervision. To overcome generative inconsistencies and the lack of multi-view capture, our pipeline introduces three core components. First, we propose a Canonical-Driven Orbital Generation (CDOG) strategy, utilizing specialized Canonical and Orbit LoRAs to normalize arbitrary input poses and synthesize reliable pseudo-multi-view images. Second, we design a Consistency-aware Deformable 3DGS (CA-3DGS) module that anchors to D-SMAL: a parametric dog prior, explicitly modeling per-view generative errors through dedicated neural deformation fields to learn accurate vertex-level displacements. Finally, to eliminate structural distortions and recover high-frequency details, we introduce a self-supervised Deformation-Conditioned Generative Repair (DCGR) module. Extensive experiments demonstrate that CORGI achieves state-of-the-art performance, generalizing seamlessly across diverse dog breeds to produce geometrically accurate, visually coherent, and fully animatable 3D assets ready for downstream applications. Project page: \url{https://dzzzby.github.io/CORGI/}},
keywords  = {3D Animal Reconstruction, Single-Image Reconstruction, 3D Gaussian Splatting, Cross-View Consistency},
copyright = {The Author(s)},
}

\begin{document}

\maketitle

\enlargethispage{-3pt}
\begin{figure}[b] \vskip -4mm
\small\renewcommand\arraystretch{1.3}
\begin{tabular}{p{80.5mm}} \toprule\\ \end{tabular}
\vskip -4.5mm \noindent \setlength{\tabcolsep}{1pt}
\begin{tabular}{p{3.5mm}p{80mm}}
$1\quad $ & School of Mathematical Sciences, University of Science and Technology of China, Hefei, 230026, China. E-mail: W. Wu, wuyx2020@mail.ustc.edu.cn; W. Li, liweile@mail.ustc.edu.cn; B. Zhu, dzzzby@mail.ustc.edu.cn; Y. Liu, lym29@mail.ustc.edu.cn\cor{}; Y. Cai, caiyoucheng@ustc.edu.cn\cor{}; X. Fu, fuxm@ustc.edu.cn ; L. Liu, lgliu@ustc.edu.cn.\\
$\dagger$ & These authors contributed equally to this work and should be considered co-first authors.\\
% $2\quad $ & Business or academic affiliation of Second B. Author, with city, post code, and country. E-mail: authorB@email.cn.\\
% $3\quad $ & Business or academic affiliation of Third C. Author, with city, post code, and country. E-mail: authorB@email.cn.\\
% &{\textcolor{blue} {(If the authors are from the same affiliation, then the same superscript should be marked on each author's name. Please provide each author's official email address)}}\\
% &\hspace{-5mm} Manuscript received: 2025-01-01; accepted: 2025-01-01\vspace{-2mm}
\end{tabular} \vspace {-3mm}
\end{figure}

\begin{figure*}[h]
    \centering
    \includegraphics[width=\textwidth, keepaspectratio]{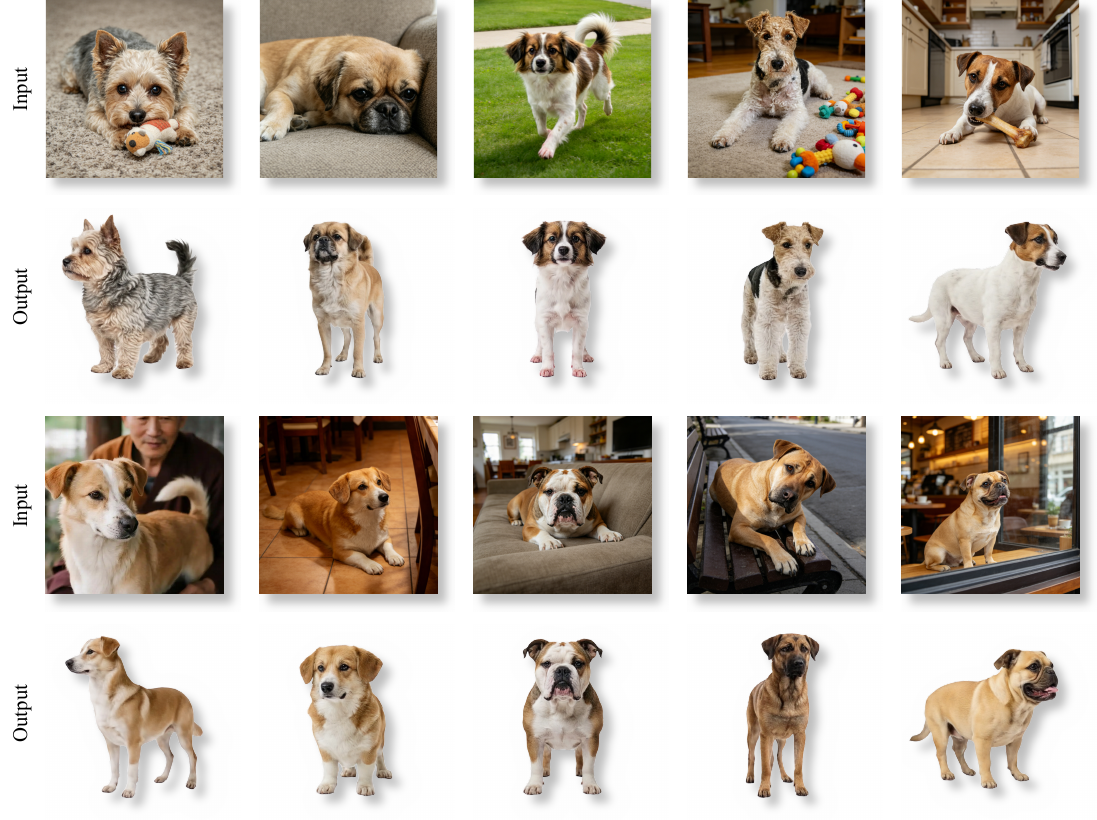}
    \caption{We propose \textbf{CORGI}, a novel framework for reconstructing a high-fidelity 3DGS dog in canonical pose from a single in-the-wild image.}
    \label{fig:teaser}
\end{figure*}

\section{Introduction}
Reconstructing the 3D shape of animals has long been a fundamental challenge in computer vision and computer graphics, supporting a wide range of applications in AR/VR and digital content creation. Among non-human species, dogs constitute a particularly compelling target because, as highly articulated quadrupeds living in close association with humans, they exhibit substantial structural variability and complex motion dynamics that continue to attract significant research interest \cite{li2025advances}. Animals are inherently non-cooperative and exhibit continuous non-rigid deformations, making calibrated multi-view capture highly impractical. Consequently, reconstructing high-fidelity 3D representations of dogs in the wild from a single input image with arbitrary pose, viewpoint, and background remains an open and important problem.

Existing methods for 3D animal reconstruction can generally be categorized into template-based and template-free approaches. Template-based approaches employ a parametric 3D template to constrain the ill-posed solution space. For instance, BITE \cite{ruegg2023bite} extends the widely adopted SMAL model \cite{zuffi20173d} into the canine domain through the proposed D-SMAL representation, thereby improving pose estimation performance, while AnimalAvatar \cite{sabathier2024animal} further enhances dynamic shape tracking from casually captured videos. Nevertheless, constrained by the limited representational capacity of the templates, template-based approaches are inherently limited in capturing subject-specific geometric details. Conversely, template-free approaches \cite{li2024learning, yao2022lassie, wu2023magicpony} aim to reconstruct articulated 3D shapes from image collections without relying on predefined templates, thereby offering greater flexibility and applicability. These methods often depend heavily on silhouette or semantic consistency, which often results in overly smooth geometries and insufficient high-frequency details when applied to single in-the-wild image.

Recently, the advent of generative 3D reconstruction has enabled new paradigms for single-view modeling. To overcome the scarcity of multi-view data for animal subjects required by traditional reconstruction methods, multi-view diffusion models \cite{liu2023zero, liu2024syncdreamer, liu2023one12345++} and video diffusion models \cite{wu2025genfusion, ren2025gen3c} generate virtual observations from a single image, which are then leveraged by reconstruction pipelines to recover the underlying 3D shape. In the domain of dogs, recent works such as DogRecon \cite{cho2025dogrecon} leverage the canine prior encoded in D-SMAL to guide multi-view image synthesis, followed by animatable 3D Gaussian Splatting (3DGS) \cite{kerbl20233d} reconstruction. Despite these promising advancements, a critical limitation persists: existing generative models frequently suffer from severe cross-view inconsistencies, leading to blurred textures and geometric artifacts during 3D optimization.

Inspired by the powerful capabilities of modern generative models, we adopt a “generation-then-reconstruction” pipeline to reconstruct high-fidelity 3D dog models from a single in-the-wild image. However, applying this paradigm to highly articulated animals introduces two major challenges. Unlike humans, it is hard to let animals cooperate with scanning or controlled multi-view capture. As a result, obtaining perfectly aligned real-world multi-view images together with corresponding ground-truth 3D models is extremely difficult, making supervised training impractical. Consequently, we must rely heavily on generative models to synthesize multi-view observations for reconstruction. However, this introduces a second challenge: due to the inherent nature of generative models, the synthesized results often exhibit inconsistent textures and geometries across views, which is undesirable for coherent 3D reconstruction.

To address these challenges, we introduce \textbf{CORGI}, a novel framework for consistency-aware 3D dog reconstruction from a single in-the-wild image. Specifically, our pipeline comprises three core components. First, we propose the \textbf{Canonical-Driven Orbital Generation (CDOG)} strategy, which transforms a single in-the-wild animal image with an arbitrary pose into a 360-degree orbital sequence anchored at a canonical standing pose for 3D reconstruction, thereby providing dense and structured multi-view observations for 3D reconstruction.  
Second, we introduce the \textbf{Consistency-Aware Deformable 3DGS (CA-3DGS)} module, which enables robust reconstruction from inconsistent pseudo-multi-view observations by recovering a base 3DGS representation following a dog prior while modeling view-specific inconsistencies with neural deformation fields.
Third, because these deformation fields are tied to the observed views and do not generalize well to unseen viewpoints, we propose the \textbf{Deformation-Conditioned Generative Repair (DCGR)} module and a novel self-supervised training strategy to repair artifacts in the reconstructed CA-3DGS and improve novel-view synthesis. 
Together, these designs enable the reconstruction of high-quality animatable 3D dog assets from a single in-the-wild image.
Extensive experiments demonstrate that our method achieves state-of-the-art reconstruction performance across the immense structural diversity of dogs; proving that CORGI generalizes seamlessly not just to Corgis, but to diverse breeds like Border Collies, while effectively supporting downstream realistic animation.

In summary, our main contributions are as follows:
\begin{itemize}
    \item We introduce \textbf{CORGI}, a novel system capable of reconstructing high-fidelity 3D dog from a single in-the wild-image, effectively eliminating the requirement for paired 3D training data.
    \item Our method addresses the severe ambiguity of monocular 3D reconstruction by using \textbf{CDOG} to expand a single image with arbitrary pose into dense observations, and propose \textbf{CA-3DGS} to recover a robust initial 3D representation from these generated views despite their cross-view inconsistencies.
    \item We introduce \textbf{DCGR} with a self-supervised learning framework that effectively repaire generative artifacts, producing geometrically accurate and visually coherent 3D models suitable for downstream animation applications.
\end{itemize}

\section{Related work}
\subsection{Animal 3D Reconstruction}
\textbf{Template-based approaches.} Inspired by the monumental success of parametric models in human body digitization, early and foundational efforts in 3D animal reconstruction heavily relied on template-based or parametric priors to constrain the severely ill-posed nature of monocular reconstruction. The pioneering SMAL model~\cite{zuffi20173d} established a skinned multi-animal linear formulation, which subsequently catalyzed the development of species-specific adaptations. Notable examples include D-SMAL~\cite{ruegg2023bite} tailored for the unique skeletal kinematics of canines, as well as hSMAL~\cite{li2021hsmal} and VAREN~\cite{zuffi2024varen} designed for equines. Building upon these robust geometric priors, a plethora of methods have been proposed to recover articulated shape and pose from single images, sparse multi-view setups, or monocular videos. These frameworks typically employ either computationally intensive optimization-based fitting or efficient feed-forward regression networks~\cite{yang2021lasr, rueegg2022barc, ruegg2023bite, Animer}. More recently, the field has witnessed a paradigm shift toward advanced rendering primitives. For instance, GART~\cite{GART} elegantly extended the template-based paradigm by replacing traditional mesh surfaces with 3D Gaussian primitives coupled with learnable skinning weights, thereby enabling highly efficient, animatable reconstruction with enhanced rendering quality.

\textbf{Template-free approaches.} Despite their robustness, template-based methods are inherently bottlenecked by the limited topological expressivity of predefined meshes, making them ill-equipped to capture subject-specific geometric variations, such as fluffy fur or distinct ear shapes across different dog breeds. To circumvent these topological constraints, a parallel and flourishing line of research explores template-free reconstruction. These methodologies typically leverage neural implicit representations, articulated neural parts, or canonical-to-posed feature decompositions to learn category-level priors directly from data. For instance, BANMo~\cite{Banmo} pioneered the disentanglement of canonical shape and non-rigid motion to learn animatable neural implicit models from monocular videos. Scaling this concept up, MagicPony~\cite{wu2023magicpony} and 3D-Fauna~\cite{li2024learning} successfully extracted category-level articulated models directly from in-the-wild image collections. Concurrently, part-based frameworks—such as LASSIE~\cite{yao2022lassie}, Hi-LASSIE~\cite{yao2023hi}, LEAPARD~\cite{LEPARD}, and ARTIC3D~\cite{ARTIC3D}—approach the problem by discovering semantic correspondences and assembling animals through self-supervised articulated neural parts. Further broadening the design space, methods like CASA~\cite{wu2022casa} and DualPM~\cite{dualpm} introduced category-agnostic skeletal reasoning and canonical point-map representations to handle extreme topological diversity.

Despite this rapid progress, reconstructing highly articulated animals from in-the-wild images remains profoundly challenging. The scarcity of large-scale, high-quality 3D ground truth paired with diverse in-the-wild images forces existing methods to rely on synthetic data, toy scans, or weak 2D supervision. Consequently, both template-based and template-free approaches often struggle to generalize across the vast structural variations of different dog breeds, suffering severe geometric degradation when confronted with complex poses, self-occlusions, and fine-scale details like fur.

\subsection{Single-Image 3D Generation}
Recovering 3D content from a single image has long been a fundamental, albeit inherently ill-posed, problem in computer vision and graphics~\cite{wang2025diffusion}. Early approaches typically formulated this as a supervised reconstruction task, training models to directly regress 3D geometry, shape parameters, or category-specific representations from a single input image~\cite{tatarchenko2019single, fu2021single, kato2019learning, li2020self, fahim2021single}. While these pioneering methods established the foundational paradigm of single-image 3D reconstruction, they were often bottlenecked by their reliance on restrictive object categories and limited geometric representations. Beyond direct reconstruction, a crucial parallel line of work introduced 3D-aware neural rendering and generative representations, which subsequently became vital priors for image-conditioned 3D generation. Methods such as GRAF~\cite{schwarz2020graf}, GIRAFFE~\cite{niemeyer2021giraffe}, StyleNeRF~\cite{gu2021stylenerf}, and EG3D~\cite{chan2022efficient} demonstrated that radiance-field-based or geometry-aware representations could synthesize view-consistent images while effectively capturing the underlying 3D structure.
 
With the proliferation of diffusion models, a highly influential line of research emerged that leverages pretrained 2D diffusion models to optimize an underlying 3D representation via diffusion guidance or score distillation. DreamFusion~\cite{poole2022dreamfusion} introduced Score Distillation Sampling (SDS) to optimize a neural radiance field using a 2D diffusion prior. Follow-up frameworks, including LucidDreamer~\cite{liang2023luciddreamer}, Magic123~\cite{qian2024magic123}, and ProlificDreamer~\cite{wang2024prolificdreamer}, further refined this approach by improving geometry initialization, optimization stability, and overall visual fidelity. Although these optimization-based methods can produce exceptionally high-quality results, they inherently require computationally expensive per-instance optimization, rendering them less suitable for scalable, feed-forward reconstruction.

To overcome the efficiency bottleneck of per-instance optimization, another major direction reformulates single-image 3D generation as a multi-view synthesis problem. Zero-1-to-3~\cite{liu2023zero} demonstrated that diffusion models could synthesize plausible novel views from a single input image, catalyzing a new pipeline that first generates sparse views and subsequently recovers 3D geometry. Later methods such as SyncDreamer~\cite{liu2024syncdreamer}, MVDream~\cite{shi2023MVD}, and Wonder3D~\cite{long2023wonder3d} significantly improved cross-view consistency and geometric coherence in the hallucinated views. Building upon this multi-view paradigm, recent Large Reconstruction Models (LRMs) have substantially advanced inference efficiency by amortizing the reconstruction process into a single feed-forward network, including LRM~\cite{hong2023lrm}, PF-LRM~\cite{wang2023pflrm}, CRM~\cite{wang2024crm}, LGM~\cite{tang2024lgm}, InstantMesh~\cite{xu2024instantmesh}, TripoSR~\cite{tochilkin2024triposr}, and One-2-3-45~\cite{liu2024one2345}.

Beyond optimization-based techniques and multi-view diffusion pipelines, the most recent frontier explores native 3D generative models aimed at modeling 3D assets directly within compact latent spaces. Representative examples include CLAY~\cite{zhang2024clay}, 3DShape2VecSet~\cite{xiang2025structured}, TRELLIS/TRELLIS-2~\cite{xiang2025native}, and Hunyuan3D~\cite{zhao2025hunyuan3d}. These architectures learn structured latent spaces over native 3D representations and perform image-conditioned generation directly within those spaces. Compared to earlier pipelines, these approaches offer superior scalability and exhibit an increasingly generalized capability for 3D generation.

Despite this sweeping and rapid progress across multiple paradigms, reconstructing highly articulated animals such as dogs from a single input image remains particularly formidable. A central bottleneck is the conspicuous absence of large-scale paired supervision between real animal images and high-quality 3D assets. In practice, much of the available 3D animal data used to train modern image-to-3D models originates from synthetic repositories, static game assets, or animation models, rather than genuine 3D captures aligned with real-world images. Consequently, even the most robust general-purpose single-image-to-3D models encounter a severe domain gap when applied to in-the-wild dog images. This discrepancy inevitably leads to drastically reduced reconstruction fidelity, poor articulation accuracy, and a loss of visual realism. By explicitly addressing these domain gaps and structural inconsistencies, our proposed framework aims to bridge the divide between generative priors and highly articulated real-world subjects.

\section{Preliminaries}
\subsection{3D Gaussian Splatting}
3D Gaussian Splatting (3DGS) \cite{kerbl20233d} has recently emerged as an efficient explicit point-based representation for novel view synthesis and 3D scene reconstruction. 3DGS models a continuous 3D scene using a collection of unstructured, anisotropic 3D Gaussians.

Each 3D Gaussian is characterized by a center position $\mathbf{\mu}$, a 3D covariance matrix $\mathbf{\Sigma}$, an opacity $\alpha$, and view-dependent color features $\mathbf{c}$, which are typically encoded using spherical harmonics (SH). The spatial contribution of a 3D Gaussian at a point $\mathbf{x}$ is defined as:
\begin{equation}
    G(\mathbf{x}) = \exp \left( -\frac{1}{2}(\mathbf{x} - \mathbf{\mu})^T \mathbf{\Sigma}^{-1} (\mathbf{x} - \mathbf{\mu}) \right).
\end{equation}

The covariance matrix $\mathbf{\Sigma}$ is decomposed into a rotation matrix $\mathbf{R}$ and a scaling matrix $\mathbf{S}$, which are parameterized by a unit quaternion and a 3D scaling vector, respectively:
\begin{equation}
    \mathbf{\Sigma} = \mathbf{R} \mathbf{S} \mathbf{S}^T \mathbf{R}^T.
\end{equation}

For efficient rendering, 3DGS employs an optimized tile-based rasterizer. Given a camera view transformation $\mathbf{W}$ and the Jacobian matrix $\mathbf{J}$ corresponding to the affine approximation of the projective transformation, the projected 2D covariance matrix $\mathbf{\Sigma}_{2D}$ is computed as follows \cite{zwicker2001ewa}:
\begin{equation}
    \mathbf{\Sigma}_{2D} = \mathbf{J} \mathbf{W} \mathbf{\Sigma} \mathbf{W}^T \mathbf{J}^T.
\end{equation}

The final color $\mathbf{C}$ of a pixel is obtained by sorting the projected Gaussians in front-to-back order and performing point-based $\alpha$-blending, thereby approximating the volume-rendering integral:
\begin{equation}
    \mathbf{C} = \sum_{i=1}^{N} c_i \alpha'_i \prod_{j=1}^{i-1} (1 - \alpha'_j),    
\label{eq:alpha-blending}
\end{equation}
where $c_i$ denotes the color of the $i$-th Gaussian encoded via spherical harmonics, $\alpha'_i$ represents the effective 2D opacity obtained by multiplying the learned opacity $\alpha_i$ with the corresponding 2D Gaussian value at the pixel location.

\subsection{Diffusion-Guided Repair}

While 3D Gaussian Splatting provides efficient and photorealistic novel-view synthesis for viewpoints close to the original camera trajectories, it inherently struggles with under-constrained regions and sparse-view settings, often producing severe artifacts such as spurious geometry, floaters, and holes \cite{wei2025gsfix3d, wu2025difix3d+}. To alleviate these limitations, recent approaches have increasingly exploited the strong generative priors of pre-trained 2D diffusion models to repair artifacts and hallucinate plausible details in unobserved regions.

The repair process is commonly formulated as a conditional image synthesis task. Given a degraded image $I_{render}$ rendered from an imperfect 3D representation, the diffusion model aims to model the conditional distribution $p(I_{gt}|I_{render})$ in order to generate a high-fidelity, artifact-free image $\hat{I}$. Mathematically, a conditional denoising network $\epsilon_\theta$ is trained to predict the noise $\epsilon$ added to the latent representation $z_t$. At timestep $t$, the network predicts the injected noise $\epsilon$ from the noisy latent representation $z_t$ by optimizing the standard denoising score-matching objective:
\begin{equation}
    \mathcal{L}_{DM} = \mathbb{E}_{z_0, \epsilon \sim \mathcal{N}(0,I), t} \left[ ||\epsilon - \epsilon_\theta(z_t, I_{render}, t)||_2^2 \right].
\end{equation}

\section{Method}
\label{sec:method}

\begin{figure*}[t]
    \centering
    \includegraphics[width=\textwidth, keepaspectratio]{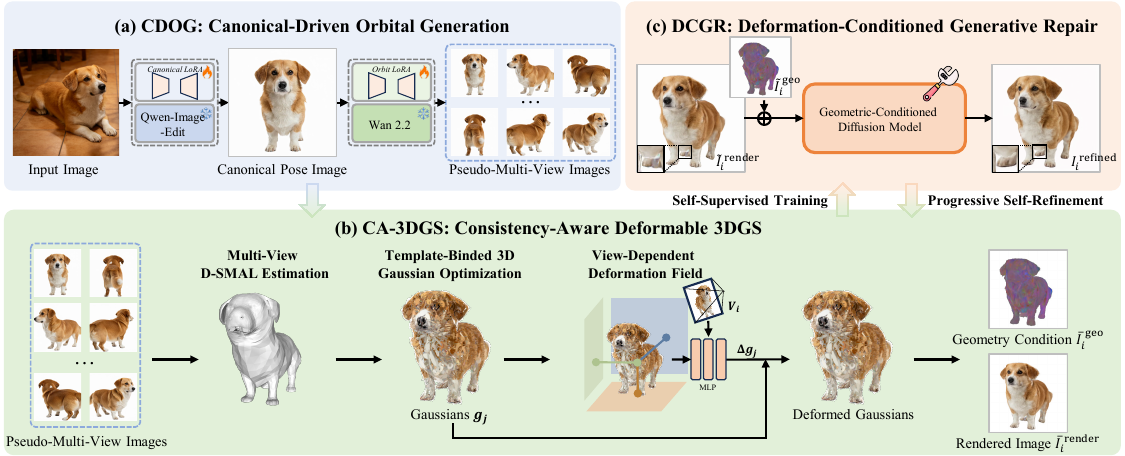}
    \caption{\textbf{Overview of the CORGI framework.} From a single input image, CORGI reconstructs a high-fidelity, animatable 3D dog without 3D supervision. (a) \textbf{CDOG} normalizes the input pose and synthesizes a reliable 360-degree pseudo-multi-view images. (b) \textbf{CA-3DGS} lifts these 2D observations into a deformable 3DGS field anchored to a D-SMAL template, using neural deformation fields to explicitly isolate view-dependent generative errors. (c) \textbf{DCGR} leverages rendered deformation maps to geometrically condition a diffusion model, rectifying residual artifacts and recovering high-frequency details in a self-supervised manner.}
    \label{fig:pipeline}
\end{figure*}

As illustrated in Figure~\ref{fig:pipeline}, we propose \textbf{CORGI}, a ``generation-then-reconstruction'' framework designed to reconstruct high-fidelity, animatable 3D dog models from a single in-the-wild input image without relying on any 3D supervision. To overcome the inherent ill-posedness and cross-view inconsistencies of single-view generation, our pipeline seamlessly integrates three core components. First, the \textbf{Canonical-Driven Orbital Generation (CDOG)} strategy employs specialized Canonical and Orbit LoRAs to normalize the arbitrary input pose and synthesize a reliable 360-degree pseudo-multi-view images. Subsequently, these 2D observations are lifted into 3D via the \textbf{Consistency-aware Deformable 3DGS (CA-3DGS)} module, which anchors 3D Gaussian primitives to a D-SMAL template and optimizes vertex-level displacements, while explicitly absorbing view-dependent generative errors through dedicated neural deformation fields. Finally, to rectify residual geometric distortions and recover high-frequency textures, we introduce the \textbf{Deformation-Conditioned Generative Repair (DCGR)} module, which leverages a 2D continuous deformation map as geometric conditioning for a pre-trained diffusion model to refine the overall 3D representation under a novel self-supervised training strategy.

\subsection{Canonical-Driven Orbital Generation} 
\label{sec:cdog} 
Acquiring perfectly aligned, multi-view captures of dynamic animals in the wild is notoriously difficult, making direct 3D supervision virtually impossible. To overcome this, we leverage an image generation model to transform an input image from an arbitrary pose and cluttered background into a canonical pose with a clean background, and subsequently employ a video generation model to synthesize a 360-degree orbit video for 3D reconstruction. A straightforward strategy to circumvent this data scarcity would be to rely heavily on test-time prompt engineering, manually adjusting text prompts conditioned on the input image to coax a pre-trained generative model into synthesizing the desired outputs. However, such heuristic prompt tuning is highly inefficient and notoriously unstable. To address this problem, we shift the paradigm from unpredictable test-time prompting to systematically fine-tuning pre-trained generative models with Low-Rank Adaptation (LoRA) \cite{hu2022lora}. Our Canonical-Driven Orbital Generation (CDOG) module decomposes the inherently ill-posed single-view reconstruction problem into two tractable, stage-wise generative processes: Canonical Pose Generation and Orbital Video Generation.

\subsubsection{Canonical Pose Generation}
The objective of this stage is to transform an input image, which captures a dog in an arbitrary pose and cluttered environment, into a canonical pose image, characterized by a canonical standing pose and a clean background.

\textbf{Dataset.} To enable consistent pose transformation, we introduce a self-supervised data curation paradigm. Utilizing the advanced generative capabilities of Qwen-Image-Edit \cite{wu2025qwenimage} guided by meticulously engineered prompts, we efficiently generate and filter a curated dataset containing 100 high-quality arbitrary-canonical pose image pairs. Each pair consists of a canine subject in a natural, arbitrary pose alongside its exact identity counterpart in a canonical standing pose with a clean background.

\textbf{Framework.} Building upon the robust zero-shot editing capabilities of Qwen-Image-Edit \cite{wu2025qwenimage} as our baseline, we formulate the pose normalization as a conditional image-to-image translation task. Using our self-supervised paired dataset, we fine-tune a \textit{Canonical LoRA} within the transformer backbone of the baseline model. This explicit fine-tuning forces the network to learn the conditional mapping from the arbitrary input to the canonical pose, effectively disentangling the subject's intrinsic identity and high-frequency textures from its original arbitrary pose. This provides a reliable spatial anchor for the subsequent orbital video generation.
\subsubsection{Orbital Video Generation} 
Given the canonical pose image, this stage synthesizes a dense sequence of multi-view observations to mimic a 360-degree camera flythrough.

\textbf{Dataset.} Due to the scarcity of high-quality orbital videos specifically featuring dogs, we deliberately assemble a specialized training dataset comprising 100 high-quality orbital videos that include a mixture of canine and \textit{non-canine} subjects. Notably, our primary objective is for the model to exclusively learn the pure geometric prior of a 360-degree azimuthal camera trajectory. Therefore, the inclusion of non-canine data provides excellent supervisory signals for camera motion without being strictly tied to the subject's semantics.

\textbf{Framework.} We inject an \textit{Orbit LoRA} into a pre-trained Image-to-Video (I2V) diffusion backbone, Wan~2.2 \cite{wan2025}, and train it exclusively on our curated orbital dataset. This training strategy successfully decouples the spatial-temporal camera motion from the semantic identity of the subject. Finally, to enforce the cyclic consistency required by the subsequent 3DGS optimization, we utilize a dual-frame conditioning mechanism, setting both the initial and terminal frames strictly to the generated canonical pose image. This forces the diffusion trajectory to synthesize a perfect closed-loop orbital video sequence, providing robust and consistent pseudo-multi-view images.

\subsection{Consistency-Aware Deformable 3DGS}
\label{sec:CA-3DGS}

Given the pseudo-multi-view images $\{I_i^{\text{gt}}\}_{i=1}^N$ generated by the CDOG module, our goal is to reconstruct a high-fidelity canonical 3DGS representation of the subject. While standard 3DGS demonstrates remarkable rendering quality and efficiency for static scenes, directly applying it to generatively synthesized multi-view data is inherently problematic. Diffusion-generated videos often exhibit view-dependent inconsistencies, texture flickering, and local geometric shifts. To address these issues, we propose the Consistency-aware Deformable 3DGS (CA-3DGS) representation. CA-3DGS binds the 3D Gaussian primitives to a parametric canine template (D-SMAL) and explicitly models generative inconsistencies through a view-dependent neural deformation field.

\subsubsection{Multi-View D-SMAL Estimation}

Unlike prior single-view methods \cite{ruegg2023bite, cho2025dogrecon}, we utilize the synthesized dense multi-view sequence for accurate template alignment. We estimate camera parameters $\{V_i\}_{i=1}^N$ using COLMAP and initialize the D-SMAL template with BITE \cite{ruegg2023bite}. We denote $\mathcal{R}_i(\cdot)$ as the differentiable renderer associated with the camera parameters of the $i$-th view. We represent the D-SMAL parameters as
$S = (\beta, \theta, \gamma, \phi, \rho)$, where $\beta$ denotes the shape parameters, $\theta$ denotes the 6D rotations of all the joints, $\gamma$ denotes the global translation, $\phi$ denotes the global orientation, and $\rho$ is a global scaling factor to accommodate dogs of varying sizes. This parameterization follows DogMo \cite{wang2025dogmo}. We denote the set of all vertices of the resulting D-SMAL template by $X$ and the 3D articulated joints by $J$. We optimize the parameters jointly across all $N$ views by minimizing the combined objective $\mathcal{L}_{\text{D-SMAL}} = \mathcal{L}_{mk} + \mathcal{L}_{kp}$. The two terms are defined as follows.

\textbf{Soft Mask Loss.} The term $\mathcal{L}_{mk}$ denotes the soft mask loss, which encourages the projected mesh to align with the multi-view silhouettes: 
$$\mathcal{L}_{mk} = \sum_{i=1}^N\|M_i-\mathcal{R}^{\text{soft}}_i(X)\|_2,$$
where $M_i$ denotes the foreground pixel set of the segmentation mask in the $i$-th view, and $\mathcal{R}^{\text{soft}}_i(X)$ denotes the soft mask rendering of of the mesh vertices $X$ in the $i$-th view as introduced in~\cite{liu2019softrasterizerdifferentiablerenderer}. Unlike a binary silhouette, the soft mask outputs a continuous probability value per pixel, indicating the likelihood of being covered by the projected mesh. This differentiability allows gradient flow through the mask loss.

\textbf{Keypoint Loss.} The term $\mathcal{L}_{kp}$ denotes the sparse keypoint loss, which constrains the articulated pose through sparse keypoint supervision:
\begin{equation}
    \mathcal{L}_{kp} = \sum_{i=1}^N \| p_i - \mathcal{P}_i(J)\|_2,
\end{equation}
where $p_i$ denotes the 2D keypoints detected by BARC \cite{rueegg2022barc}, and $\mathcal{P}_i(J)$ represents the projection of the corresponding 3D joints.

\subsubsection{Template-Binded 3D Gaussian Optimization}

To maintain topological consistency and facilitate downstream animation, we explicitly attach the 3D Gaussian primitives to the surface of the optimized D-SMAL mesh. Inspired by \cite{jiang2025uv, gao2024real}, we sample Gaussians uniformly over the template surface within its 2D UV parameterization space. Let $\mu^{2D}$ denote the UV coordinates of a Gaussian. Its base 3D position $U(X, \mu^{2D})$ is computed through barycentric interpolation of the mesh vertices $X$. To capture subject-specific non-rigid details (e.g., fur) that cannot be represented by the base template, we introduce a learnable scalar $\tau$ that displaces the Gaussian along the interpolated surface normal $\mathbf{n}$. The final position of the $j$-th Gaussian is formulated as:
\begin{equation}
    \mu_j = U(X, \mu^{2D}_j) + \tau_j \mathbf{n}_j.
\label{eq:anchored-3dgs-position}
\end{equation}
Crucially, in addition to the Gaussian attributes, the underlying 3D vertices of the D-SMAL mesh are jointly optimized as learnable parameters to better fit subject-specific geometry.

\subsubsection{View-Dependent Deformation Field}

Although the orbital sequence provides dense 360-degree coverage, the generated pseudo-multi-view images exhibit local cross-view inconsistencies. Forcing a static 3DGS model to fit these inconsistent observations leads to severe texture blurring and floating artifacts. To disentangle the underlying canonical geometry from the view-dependent inconsistencies introduced by the generative model, we represent the scene using a base 3DGS model $\mathcal{G}$ together with a view-dependent neural deformation field $\mathcal{F}_\theta$. In this way, $\mathcal{G}$ captures the shared, view-consistent geometry, while $\mathcal{F}_\theta$ accounts for the view-dependent variations in the Gaussian attributes induced by the inconsistent observations.

Here we denote the base model as $\mathcal{G}=\{g_j\}_{j=1}^K$, where each Gaussian is parameterized by $g_j = (\mu_j, s_j, r_j, \alpha_j, c_j)$. Specifically, $\mu_j$ denotes the position defined in Eq.~\eqref{eq:anchored-3dgs-position}, $s_j$ and $r_j$ represent the scale and rotation of the $j$-th Gaussian respectively, $\alpha_j$ and $c_j$ are the opacity and color mentioned in Eq.~\eqref{eq:alpha-blending}.

The deformation field $\mathcal{F}_\theta$ is parameterized by an efficient triplane representation \cite{chan2022efficient} for spatial feature encoding and a lightweight Multi-Layer Perceptron (MLP) for residual deformation decoding. Given a camera view $V_i$, the deformation field $\mathcal{F}_\theta$ takes the position $\mu_j$ of the$j$-th Gaussian as input and predicts a residual offset for its attributes:
\begin{equation}
    \Delta g_{j,i} = \mathcal{F}_\theta(V_i, \mu_j).
\end{equation}

The deformed attributes used for rendering the $i$-th view are updated as:
\begin{equation}
    g'_{j,i} = g_j + \Delta g_{j,i}.
\label{eq:bias-plus}
\end{equation}
This design effectively represents the inconsistencies by the neural deformation field while preserving a clean 3D representation shared across all views.

For $i$-th view, the image is rendered from the deformed attributes $\{g'_{j,i}\}_{j=1}^K$ using the standard differentiable 3DGS rasterizer. We jointly optimize the mesh vertices $X$, the Gaussian attributes $\{\tau_j,s_j, r_j,\alpha_j,c_j\}_{j=1}^K$, and parameters of the deformation field $F_{\theta}$ using the following objective:
\begin{equation}
    \mathcal{L} = \mathcal{L}_\text{image} + \lambda_\text{arap} \mathcal{L}_\text{arap} + \lambda_{\text{reg}} \mathcal{L}_{\text{reg}},
\end{equation}
where each loss is defined as follow. Firstly, $\mathcal{L}_\text{image}$ consists of $L_1$ loss and Structural Similarity Index Measure (SSIM) to enforce pixel-wise and perceptual consistency with respect to $\{I_i^{\text{gt}}\}_{i=1}^N$. Secondly, $\mathcal{L}_\text{arap}$ denotes As-Rigid-As-Possible (ARAP) energy \cite{igarashi2005rigid}, which encourages local rigidity by constraining each vertex neighborhood to preserve its original shape up to a local rotation. Specifically, it is computed by penalizing discrepancies between the deformed edge vectors and the corresponding template edge vectors after optimal local rotation alignment. Finally, $\mathcal{L}_{\text{reg}}$ regularizes the deformation field to encourage sparse and smooth corrections:
\begin{equation}
    \mathcal{L}_{reg} = \frac{1}{NK} \sum_i^N \sum_j^K \|\Delta g_{j,i}\|_2^2 + \lambda_\text{grad}\mathcal{L}_{\text{grad}},
\end{equation}
where $K$ is the total number of Gaussian primitives. The first term penalizes the magnitude of the deformations to suppress large and unstable deformations. The second term, $\mathcal{L}_{\text{grad}}$, penalizes spatial variations in the deformation residuals, whose gradients are approximated using finite differences on the triplane grid, thereby explicitly enforcing spatial smoothness.

\subsection{Deformation-Conditioned Generative Repair}
\label{sec:dcgr}

Although the view-dependent deformation field $\mathcal{F}_\theta$ absorbs per-view inconsistencies during optimization, its predictions are inherently tied to the input view. If we render each view $V_i$ using its corresponding deformation, the resulting 3DGS would reproduce the view inconsistencies present in the generated inputs. Therefore, instead of applying view-specific deformations at inference time, we freeze the neural deformation field to a single reference view (typically the first view $V_0$) and use the resulting deformed geometry for rendering all novel views. Ideally, multi-view consistent images $\{\tilde{I}_i^{\text{render}}\}_{i=1}^N$ should be rendered by the differentiable renderer $\mathcal{R}_i$, where
\begin{equation}
    \tilde{I}_i^{\text{render}} = \mathcal{R}_i \Big( \big\{ g_j + \mathcal{F}_\theta(V_0, \mu_j) \big\}_{j=1}^K \Big).
    \label{eq:render}
\end{equation}
However, because the deformation field $\mathcal{F}_\theta(V_0, \cdot)$ is inherently designed to fit the specific generative flaws of the reference view $V_0$, forcing it to adapt to a drastically different view $V_i$ inevitably causes severe degradation, such as geometric tearing and texture blurring. To mitigate these artifacts, we introduce the Deformation-Conditioned Generative Repair (DCGR) module, which leverages a custom self-supervised learning paradigm to restore degraded image to high-fidelity, artifact-free results.

\begin{figure}[h]
    \centering
    \includegraphics[width=\columnwidth, keepaspectratio]{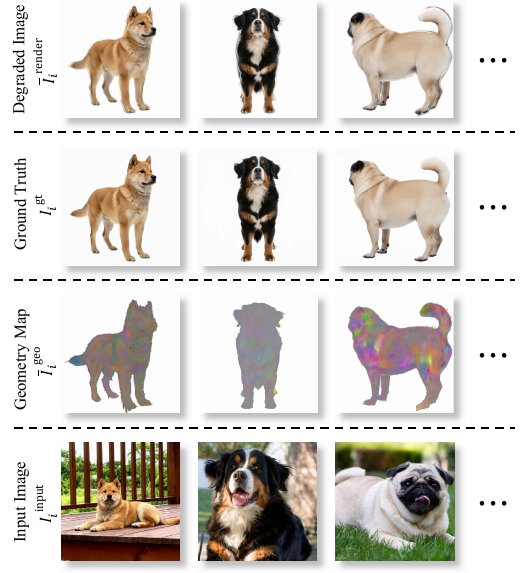}
    \caption{\textbf{Creation of Training Dataset.} From top to bottom, each row respectively represents: the degraded image $\bar{I}_i^{\text{render}}$, the ground truth ${I}_i^{\text{gt}}$, the geometry condition $\bar{I}_i^{\text{geo}}$, and the input image ${I}_i^{\text{input}}$ as an appearance reference.}
    \label{fig:dataset}
\end{figure}
\subsubsection{Self-Supervised Dataset Creation}
To train the DCGR module for correcting generative artifacts under the perspective-locked rendering formulation in Eq.~\ref{eq:render}, we construct a self-supervised dataset that captures the characteristic degradations produced by the initial 3DGS model without requiring clean 3D ground truth.

To ensure sufficient diversity of real images in the training dataset, we collect a large number of in-the-wild dog images from the Internet, denoted as $I^{\text{input}}$. We then apply the CDOG module to generate pseudo multi-view images $\{I_i^{\text{gt}}\}_{i=1}^{N}$, which serve as pseudo ground truths, and optimize the CA-3DGS model along with its view-dependent deformation field $\mathcal{F}_{\theta}$. Hereafter, $\{g_j\}_{j=1}^{K}$ denotes the attributes of the optimized Gaussians. Because the CA-3DGS model and the deformation field $\mathcal{F}_{\theta}$ are optimized from inconsistent multi-view images, the resulting renderings naturally contain artifacts. For each camera view $V_i$, we render the degraded image $\bar{I}_i^{\text{render}}$ by using $V_i$ as the input to $\mathcal{F}_{\theta}$:
\begin{equation}
    \bar{I}_i^{\text{render}} = \mathcal{R}_i \Big( \big\{ g_j + \mathcal{F}_\theta(V_i, \mu_j)\big\}_{j=1}^K \Big).
\end{equation}
Notably, $\{\bar{I}_i^{\text{render}}\}_{i=1}^{N}$ are multi-view inconsistent renderings that serve as the degraded images for DCGR, while the corresponding pseudo ground truths $\{I_i^{\text{gt}}\}_{i=1}^{N}$ are used as supervision targets.

Meanwhile, to provide explicit structural guidance regarding these geometric artifacts, we project the view-specific displacement vectors onto the image plane to obtain a geometry map:
\begin{equation}
    \bar{I}_i^{\text{geo}} = \mathcal{R}_i \Big( \big\{ \mu'_{j,i}, s'_{j,i}, r'_{j,i}, \alpha'_{j,i}, \Delta\mu_{j,i} \big\}_{j=1}^K \Big),
\end{equation}
where $\mu'_{j,i}, s'_{j,i}, r'_{j,i}$, and $\alpha'_{j,i}$ are defined in Eq.~\eqref{eq:bias-plus}.

Consequently, for each view $i$, we construct a self-supervised training quadruplet
$\{\bar{I}_i^{\text{render}}, I_i^{\text{gt}}, \bar{I}_i^{\text{geo}}, I^{\text{input}}\}$,
where $\bar{I}_i^{\text{render}}$ is the degraded image, $I_i^{\text{gt}}$ is the pseudo ground truth, the geometry map $\bar{I}_i^{\text{geo}}$ serves as the geometric condition, and the input image $I^{\text{input}}$ provides the appearance reference.

\subsubsection{Network Framework and Progressive Self-Refinement}
\textbf{Geometric-Conditioned Diffusion Model.} 
We build our repair framework upon the single-step diffusion architecture of DiFix3D+ \cite{wu2025difix3d+}, formulating the artifact removal as a geometrically-conditioned translation task. To adapt the generic diffusion prior to our canine reconstruction pipeline, we freeze the base weights and inject a lightweight LoRA module into the backbone.

\emph{(1) Training.} During training, only the inserted LoRA parameters are optimized while the pretrained DiFix3D+ backbone remains frozen. Given a self-supervised training quadruplet
$\{\bar{I}_i^{\text{render}}, I_i^{\text{gt}}, \bar{I}_i^{\text{geo}}, I^{\text{input}}\}$,
the degraded image and the geometry map are concatenated as the structural condition, whereas the input image provides the appearance guidance. The diffusion model is trained to reconstruct the corresponding pseudo ground truth $I_i^{\text{gt}}$ using the standard single-step denoising objective adopted in DiFix3D+:
\begin{equation}
\mathcal{L}_{\text{diff}}
=
\mathbb{E}_{z_t,\epsilon}
\left[
\|
\epsilon-
\epsilon_\theta(z_t,c)
\|_2^2
\right],
\end{equation}
where $z_t$ denotes the noisy latent at diffusion step $t$, $\epsilon$ is the sampled Gaussian noise, and
\begin{equation}
c=
\{\bar{I}_i^{\text{render}},\bar{I}_i^{\text{geo}},I^{\text{input}}\}
\end{equation}
represents the geometric and appearance conditions. Since only the LoRA layers are optimized, the pretrained diffusion prior is efficiently adapted to geometry-aware artifact removal while preserving its powerful image restoration capability.

\emph{(2) Inference.} During inference, given a rendered image
$\tilde{I}_{i}^{\text{render}}$
from the current CA-3DGS model, we first generate its corresponding geometry map
$\tilde{I}_{i}^{\text{geo}}$
by projecting the view-dependent Gaussian displacements onto the image plane. The rendered image, geometry map, and appearance reference $I^{\text{input}}$ are then jointly fed into the LoRA-adapted diffusion model to directly predict a repaired image
$\tilde{I}_{i}^{\text{refined}}$
through a single denoising step. Benefiting from the complementary geometric and appearance guidance, the repaired result preserves the identity and fine-grained texture of the input dog while effectively eliminating rendering artifacts.

\textbf{Progressive Self-Refinement.} 
To propagate these 2D image enhancements back into the 3D space, we wrap the LoRA-adapted DiFix3D+ network within an progressive self-refinement loop. Starting from the initial representation $\mathcal{G}^0$, which is the optimized CA-3DGS obtained in Section~\ref{sec:CA-3DGS}, each refinement round $t$ executes three sequential operations: 

\emph{(1) Rendering.} We render the multi-view images $\{\tilde{I}_{i,t}^{\text{render}}\}_{i=1}^N$ from the current model $\mathcal{G}^t$ according to Eq.\ref{eq:render}, while fixing the deformation field to the reference view $V_0$ to enforce global consistency. 

\emph{(2) Generative Repair.} For each camera view $V_i$,  the rendered image $\tilde{I}_{i,t}^{\text{render}}$, its geometry map $\tilde{I}_{i,t}^{\text{geo}}$, and the appearance reference $I_{\text{input}}$ are jointly fed into our geometric-conditioned diffusion model to produce a refined, artifact-free observation $I_{i,t}^{\text{refined}}$. 

\emph{(3) Self-Refinement.} The original pseudo-multi-view images generated by CDOG are replaced with these repaired images to serve as new pseudo ground truths, and the 3DGS parameters are re-optimized using the losses defined in Section~\ref{sec:CA-3DGS} to yield the updated model $\mathcal{G}^{t+1}$. In practice, we find that performing the self-refinement loop up to $t=3$ is sufficient to obtain a high-fidelity and fully consistent 3DGS model of the dog.

\section{Experiments}
\label{sec:experiments}

\subsection{Experiment Setup}

\,\,\,\, \textbf{Implementation details.}
The proposed CORGI framework is implemented in PyTorch. For the Canonical-Driven Orbital Generation (CDOG) module, we generate an 80-frame surrounding video to provide dense pseudo-multi-view images for the subsequent reconstruction stage. The Consistency-Aware Deformable 3DGS (CA-3DGS) module is optimized for a total of 30,000 iterations. The regularization weights for the deformation field are empirically set to $\lambda_\text{arap} = 0.01$, $\lambda_\text{reg} = 0.01$, and $\lambda_\text{grad} = 1$. For the Deformation-Conditioned Generative Repair (DCGR) module, the progressive self-refinement loop is executed for $3$ rounds, with each round undergoing 8,000 iterations of re-optimization. All experiments are conducted on a single NVIDIA RTX 4090 (48GB) GPU.

\textbf{Datasets.}
To rigorously evaluate our framework across both constrained and unconstrained scenarios, we curate two distinct evaluation datasets: (1) \textbf{Dog Synthetic Dataset:} This dataset comprises 10 high-quality, fully rigged 3D canine models. For each model, we render 24 views along a 360-degree orbital trajectory. The first frame strictly serves as the monocular input, while the remaining 23 views are reserved exclusively as ground truth (GT) to quantitatively evaluate multi-view rendering accuracy and geometric fidelity. (2) \textbf{Dog Wild Dataset:} To assess in-the-wild generalization capability, we collect 150 real-world dog images sourced from the internet. This dataset covers a vast spectrum of dog breeds, arbitrary poses, varying illuminations, and cluttered backgrounds. Since no 3D ground truth is available for real images, this dataset is used exclusively to evaluate single-view 3D generation quality and robustness.

\textbf{Metrics.}
Due to the generative nature of our pipeline, we utilize both reference-based and non-reference metrics to provide a comprehensive and human-perception-aligned evaluation: 
(1) \textbf{Reference-based Metrics:} Evaluated on the Dog Synthetic Dataset, we measure multi-view consistency and reconstruction accuracy against the ground truth using PSNR, SSIM, and LPIPS~\cite{zhang2018unreasonable}. Specifically, PSNR evaluates the absolute pixel-level reconstruction error. SSIM quantifies the structural and luminance fidelity of the generated geometries. LPIPS assesses deep feature-level perceptual similarity, which aligns more closely with human visual judgment than traditional pixel-wise metrics. (2) \textbf{Non-reference Metrics:} Evaluated on the Dog Wild Dataset, we employ widely adopted perceptual metrics including FID~\cite{heusel2018ganstrainedtimescaleupdate} and  NIQE~\cite{zhang2015feature} to assess rendering realism without ground truth. Specifically, FID quantifies the distribution distance between the rendered images and the real data domain, reflecting overall generative fidelity. NIQE measures the naturalness of the synthesized views based on statistical deviations from natural scene models. Furthermore, because preserving the original subject's identity is crucial, we follow prior work~\cite{tang2024dreamgaussian} and calculate the CLIP-cosine similarity between the input image and the rendered novel views to rigorously evaluate semantic and identity consistency.

\subsection{Baselines}

\begin{table*}[t]
\centering
\caption{\textbf{Quantitative comparison of single-image 3D dog reconstruction.} CORGI achieves the overall best performance among all baselines across both reference-based and non-reference metrics. Bold indicates the best result and underlines indicates the second-best.}
\label{tab:quantitative}
\setlength{\tabcolsep}{3.8mm}
\begin{tabular}{l | c c c c | c c c}
\toprule
& \multicolumn{4}{c|}{Dog Synthetic Dataset} & \multicolumn{3}{c}{Dog Wild Dataset} \\
Method & PSNR $\uparrow$ & SSIM $\uparrow$ & LPIPS $\downarrow$ & CLIP $\uparrow$ & FID $\downarrow$ & NIQE $\downarrow$ & CLIP $\uparrow$ \\
\midrule
DreamGaussian & 13.795 & 0.827 & 0.252 & 0.880 & 87.900 & 8.363 & 0.733 \\
GenFusion & 8.546 & 0.670 & 0.481 & 0.751 & 241.330 & 8.780 & 0.642 \\
SyncDreamer & 11.668 & 0.784 & 0.348 & 0.776 & 198.621 & 10.442 & 0.668 \\
Ar-1-to-3 & 12.600 & 0.824 & 0.266 & 0.843 & 84.990 & \underline{7.421} & 0.746 \\
SVC & 11.818 & 0.815 & 0.279 & 0.862 & 63.250 & 8.228 & 0.748 \\
HunyuanWorld-Voyager & 11.532 & 0.674 & 0.445 & 0.708 & 71.670 & 7.853 & 0.701 \\
Hunyuan3D 2.0 & 12.511 & 0.816 & 0.248 & \underline{0.893} & 67.850 & 7.922 & 0.749 \\
Trellis2 & 13.564 & \textbf{0.841} & \textbf{0.228} & 0.877 & \underline{44.301} & 9.040 & \underline{0.771} \\
\midrule
GART & \underline{13.799} & 0.814 & 0.266 & 0.824 & 74.218 & 8.738 & 0.743 \\
BANMo & 13.749 & 0.811 & 0.262 & 0.854 & 99.654 & 9.409 & 0.718 \\
\midrule
CORGI & \textbf{14.309} & \underline{0.836} & \underline{0.236} & \textbf{0.904} & \textbf{13.477} & \textbf{6.515} & \textbf{0.800} \\
\bottomrule
\end{tabular}
\end{table*}

To demonstrate the superiority of CORGI, we compare our framework against a comprehensive suite of state-of-the-art baselines. These baselines encompass a wide spectrum of paradigms, ranging from regression-based techniques to diffusion-based generative approaches.

\textbf{Single-Image 3D Generation Methods.}
We select prominent baselines representing four distinct technical trajectories within the image-to-3D domain:
\begin{itemize}
    \item \textit{Image-to-Gaussian:} DreamGaussian~\cite{tang2024dreamgaussian} and GenFusion~\cite{wu2025genfusion}.
    \item \textit{Multi-view Diffusion:} SyncDreamer~\cite{liu2024syncdreamer} and Ar-1-to-3~\cite{zhang2025ar}.
    \item \textit{Video Diffusion:} SVC~\cite{zhou2025stable}, and HunyuanWorld-Voyager~\cite{huang2025voyager}. Since these methods output temporal video sequences rather than explicit 3D assets, we employ a vanilla 3DGS optimization on their 80-frame generated multi-view videos to obtain the final 3D representation for a fair structural comparison.
    \item \textit{Native 3D Generation:} Hunyuan3D 2.0~\cite{zhao2025hunyuan3d} and Trellis2~\cite{xiang2025native}.
\end{itemize}

It is crucial to note that directly feeding in-the-wild, arbitrarily posed dog images into these general-purpose baselines yields heavily distorted, structurally collapsed, or semantically incorrect geometries. Therefore, to ensure the most competitive and fair comparison possible, we utilize the canonical pose image obtained by our CDOG module as the standardized input for all aforementioned single-image baselines.

\textbf{Multi-Image 3D Reconstruction Methods.}
To further isolate and validate the specific contributions of our CA-3DGS and DCGR modules in handling dynamic/inconsistent multi-view data, we compare against robust video-driven animal reconstruction frameworks: GART~\cite{GART} and BANMo~\cite{Banmo}. For these baselines, instead of a single image, we provide the full 80-frame, 360-degree orbital video generated by our CDOG module as their input.

\subsection{Comparison}
\label{sec:comparison}

\begin{figure*}[!p]
\centering
\includegraphics[width=\textwidth]{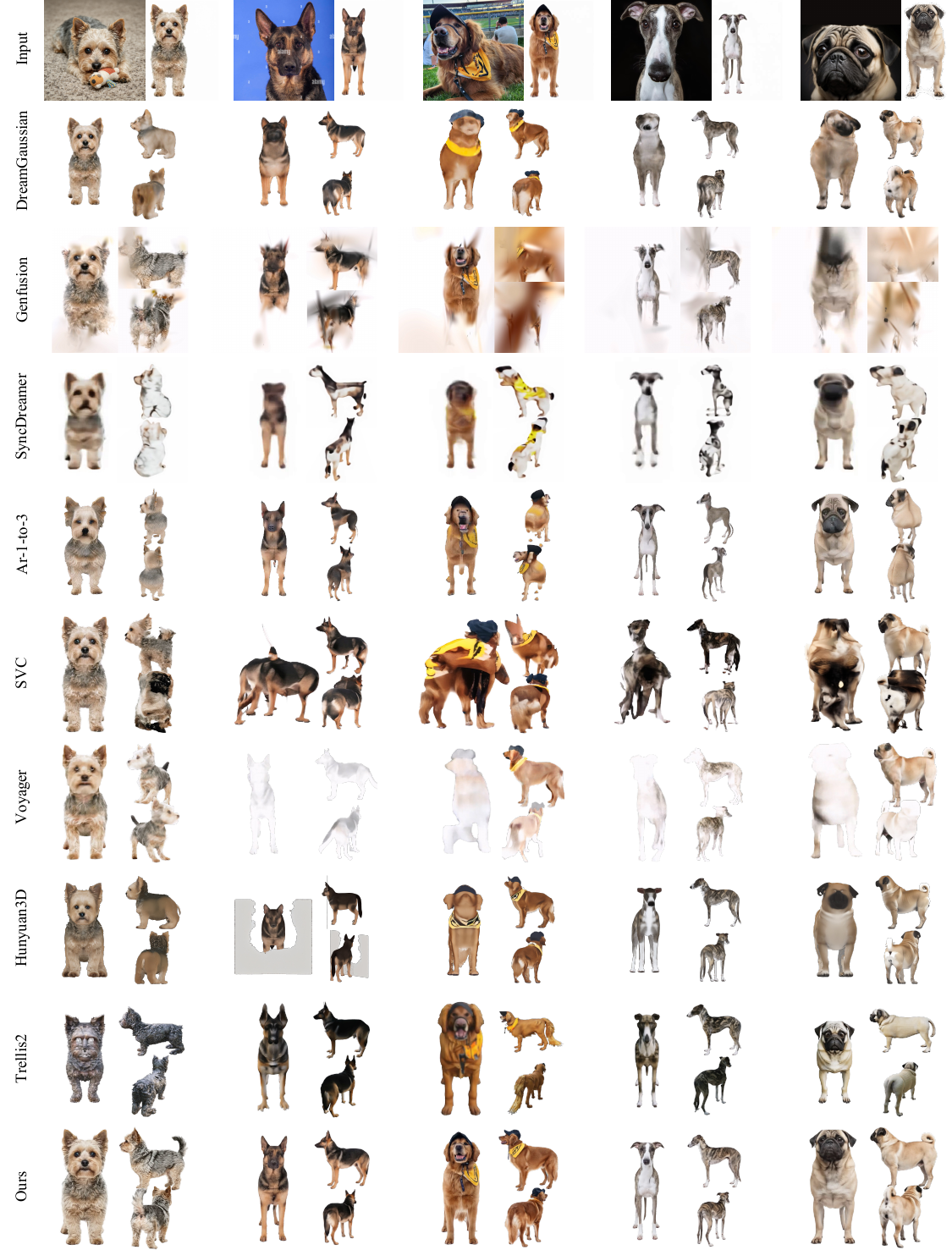}
\caption{\textbf{Qualitative comparison on in-the-wild dog images.} Compared to baselines that suffer from, blurry textures, or structural collapse, CORGI reconstructs high-fidelity, geometrically accurate 3D models with crisp fur details and strict multi-view consistency.}
\label{fig:comparison}
\end{figure*}

\textbf{Quantitative Evaluation.} 
The quantitative comparisons against all baselines on both the Dog Synthetic Dataset and the Dog Wild Dataset are summarized in Table~\ref{tab:quantitative}. Our proposed CORGI compellingly demonstrates state-of-the-art performance across all evaluation metrics. 

On the \textbf{Dog Synthetic Dataset}, CORGI achieves the highest PSNR and second highest SSIM scores, significantly outperforming Image-to-Gaussian methods (e.g., DreamGaussian~\cite{tang2024dreamgaussian}) and Multi-view Diffusion models (e.g., SyncDreamer~\cite{liu2024syncdreamer}). This superiority indicates that our Canonical-Driven Orbital Generation (CDOG) and Consistency-Aware Deformable 3DGS (CA-3DGS) modules successfully establish a rigorous 3D geometric structure rather than merely hallucinating independent 2D views. Nevertheless, CORGI also achieves the second lowest LPIPS score, proving its exceptional capability in preserving high-frequency textural details. Notably, Trellis2~\cite{xiang2025native} achieves the best SSIM and LPIPS scores, which we attribute to the nature of the synthetic data: the mesh-based geometry and smooth textures of the Dog Synthetic Dataset are inherently well-suited to Trellis2's native 3D latent representation.

On the \textbf{Dog Wild Dataset}, the non-reference metrics further highlight our method's robustness to in-the-wild domain gaps. General-purpose generative models often struggle with the complex articulation and non-rigid deformations of real-world canines. In contrast, CORGI attains the best FID and NIQE scores, reflecting the photorealism and naturalness of our rendered views. Most importantly, our method achieves a remarkable margin in the CLIP cosine similarity score. This proves that, unlike Native 3D models that frequently collapse into generating a ``generic'' dog due to template bias, our framework strictly preserves the subject-specific identity, fur patterns, and structural nuances of the input image. Furthermore, compared to video-driven multi-image baselines (GART~\cite{GART} and BANMo~\cite{Banmo}), which severely degrade when optimizing directly on our pseudo-multi-view images due to generative flickering, our method robustly absorbs these inconsistencies via the view-dependent deformation field, yielding significantly higher quantitative fidelity.

\textbf{Qualitative Evaluation.} 
The visual comparisons between CORGI and representative state-of-the-art baselines are illustrated in Figure~\ref{fig:comparison}. The qualitative results align perfectly with our quantitative findings, explicitly revealing the inherent bottlenecks of existing paradigms when applied to highly articulated animals in the wild.

As observed in Figure~\ref{fig:comparison}, \textit{Image-to-Gaussian} and \textit{Multi-view Diffusion} methods frequently suffer from ambiguity and blurry problem. Because they lack explicit structural priors for quadrupeds, they often produce geometrically collapsed bodies, missing limbs, or floaters in occluded regions. \textit{Video Diffusion} baselines (e.g., HunyuanWorld-Voyager~\cite{huang2025voyager}), while offering better temporal smoothness, lack strict 3D multi-view constraints. When their generated 80-frame videos are lifted to vanilla 3DGS, the inherent generative inconsistencies inevitably result in torn geometries and blurry, ghosting textures. 

\textit{Native 3D Generation} models (e.g., Hunyuan3D 2.0~\cite{zhao2025hunyuan3d} and Trellis2~\cite{xiang2025native}) successfully maintain 3D consistency but exhibit a profound domain gap. Trained predominantly on synthetic or rigid game assets, they struggle to capture the complex kinematics of real dogs, often outputting rigid, toy-like geometries with completely smoothed-out fur details, failing to preserve the unique identity of the input image.

% --- Placeholder for Animation Figure ---
\begin{figure*}[h]
\centering
\includegraphics[width=\textwidth, keepaspectratio]{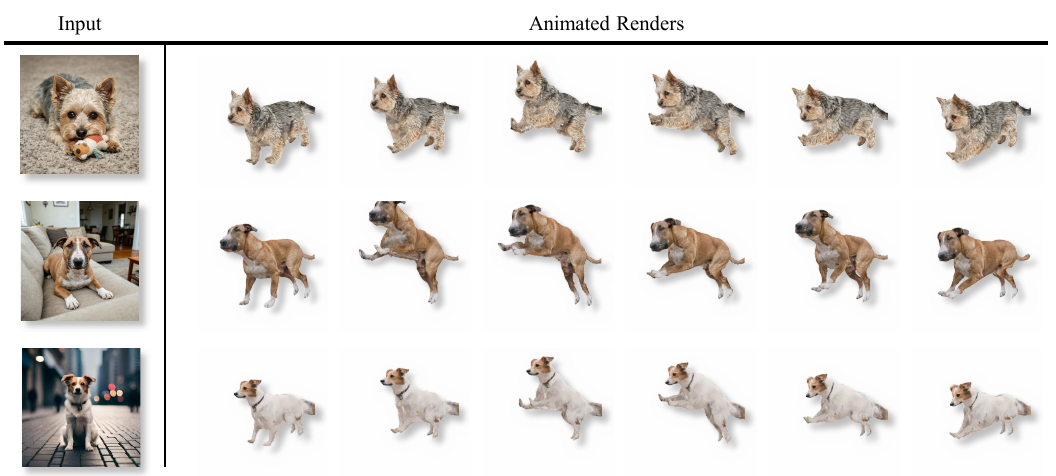} 
\caption{\textbf{Pose-driven animation results.} Driven by novel D-SMAL pose sequences, our reconstructed 3D models seamlessly perform complex motions, while strictly preserving subject-specific identities, high-frequency textures, and structural integrity without geometric tearing.}
\label{fig:animation}
\end{figure*}

\subsection{Applications}
\label{sec:applications}

CORGI explicitly overcomes these limitations. Empowered by the D-SMAL prior within the CA-3DGS module and the progressive self-refinement of the DCGR module, our method synthesizes topologically intact, structurally coherent, and highly detailed 3D canines. Even under extreme poses, severe self-occlusions, and cluttered backgrounds, CORGI accurately reconstructs delicate anatomical structures (such as thin legs, ears, and snouts) alongside crisp, high-frequency fur textures, seamlessly bridging the gap between 2D generative priors and 3D animatable articulation.

A fundamental advantage of our generation-then-reconstruction paradigm is that it yields an inherently animatable 3D representation, directly bridging the gap between unconstrained single-image generation and downstream digital content creation. We demonstrate this practical utility through high-fidelity pose-driven animation.

\textbf{Pose-Driven Animation.} 
The final reconstructed asset from our CA-3DGS module consists of an optimized 3D Gaussian field explicitly anchored to the parametric D-SMAL template. Although our initial CDOG stage synthesizes a standardized standing pose, the individually optimized subject may still exhibit slight kinematic deviations from a strict mathematical rest pose. To enable flawless and artifact-free animation, we first map the optimized Gaussians into a true canonical rest space via reverse Linear Blend Skinning (LBS). Specifically, for each Gaussian primitive bound to the mesh surface, we invert the local skinning transformations using the originally estimated pose parameters to extract its absolute rest-pose attributes. 

Once normalized into this rigorous canonical space, the 3D dog model can be seamlessly driven by arbitrary target pose sequences. By applying forward LBS parameterized by novel D-SMAL kinematic poses, we dynamically articulate the canonical Gaussians into new poses. As illustrated in Figure~\ref{fig:animation}, CORGI produces highly realistic and fluid animated sequences. Crucially, because the underlying geometry is topologically consistent and thoroughly refined by our DCGR module, the animated results exhibit zero structural tearing or floating artifacts. The model successfully preserves complex geometric nuances and high-frequency texture details (e.g., subject-specific fur patterns) even under extreme articulations and challenging novel viewpoints.

\subsection{Ablation Study}
\label{sec:ablation}

\begin{table*}[h]
\centering
\caption{\textbf{Quantitative Ablation Study.} We incrementally add our proposed modules to the baseline. The Full Model achieves the best performance across all perceptual metrics.}
\label{tab:ablation}
\begin{tabular}{l | c c c c | c c c }
\toprule
& \multicolumn{4}{c|}{Components} & \multicolumn{3}{c}{Dog Wild Dataset} \\
\cmidrule(lr){2-5} \cmidrule(lr){6-8}
Config & Def. Field & DiFix3D Prior & LoRA Fine-tune & Refine. Loop & FID $\downarrow$ & NIQE $\downarrow$ & CLIP $\uparrow$ \\
\midrule
Baseline & & & & & 26.515 & 7.719 & 0.761  \\
Model A  & \checkmark & & & & 25.934 & 8.201 & 0.781  \\
Model B  & \checkmark & \checkmark & & & 22.156 & 7.010 & 0.779  \\
Model C  & \checkmark & \checkmark & \checkmark & & 20.413 & 6.502 & 0.788  \\
\midrule
\textbf{Full Model} & \checkmark & \checkmark & \checkmark & \checkmark & \textbf{19.703} & \textbf{5.754} & \textbf{0.795}  \\
\bottomrule
\end{tabular}
\end{table*}

\begin{figure}[h]
\centering
\includegraphics[width=\columnwidth, keepaspectratio]{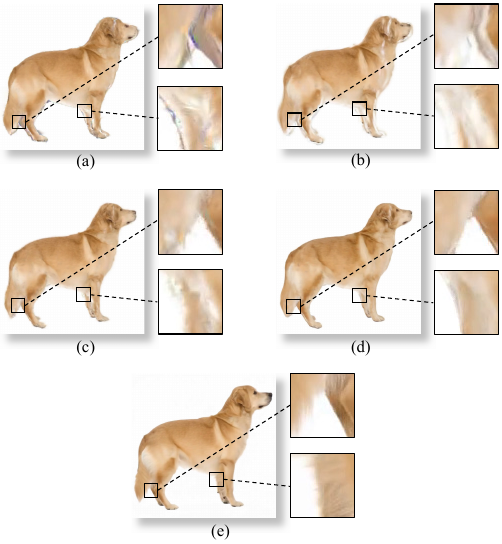} 
\caption{\textbf{Qualitative Ablation Study.} (a) \textbf{Baseline}, (b) \textbf{Model A}, (c) \textbf{Model B}, (d) \textbf{Model C}, (e) \textbf{Full Model}}
\label{fig:ablation}
\end{figure}

To rigorously validate the efficacy of our core algorithmic designs, we conduct an incremental ablation study on the Dog Wild Dataset. We define a baseline model and systematically integrate our proposed components to analyze their distinct contributions. The quantitative results across all configurations are summarized in Table~\ref{tab:ablation}.

Specifically, our evaluated configurations and their corresponding table abbreviations are defined as follows:
\begin{itemize}
    \item \textbf{Baseline:} A vanilla 3DGS optimization applied directly to the pseudo-multi-view images generated by the CDOG module.
    \item \textbf{Model A:} Baseline + View-Dependent Deformation Field (Def. Field).
    \item \textbf{Model B:} Model A + off-the-shelf DiFix3D~\cite{wu2025difix3d+} (DiFix3D Prior).
    \item \textbf{Model C:} Model A + DiFix3D with LoRA fine-tuning and geometry conditioning (LoRA Fine-tune).
    \item \textbf{Full Model:} Model C integrated with the progressive self-refinement loop for three iterations (Refine. Loop).
\end{itemize}

\textbf{Effectiveness of the Deformation Field.} 
As shown in Figure~\ref{fig:ablation}, the \textbf{Baseline} produces severe artifacts and prominent structural ghosting, particularly along the body contours. This occurs because the diffusion-generated video inherently contains multi-view inconsistencies. Forcing a static, canonical Gaussian set to strictly satisfy all of these conflicting observations simultaneously leads to severe optimization conflicts. By integrating the neural deformation field, \textbf{Model A} explicitly models these cross-view discrepancies via view-specific residual adjustments, effectively disentangling the canonical geometry from generative noise. However, forcing this deformation field to generalize to a drastically different, perspective-locked view inevitably causes structural degradation, necessitating our subsequent repair modules.

\textbf{Effectiveness of LoRA Fine-Tuning.} 
As shown in Figure~\ref{fig:ablation}, \textbf{Model B} attempts to repair the aforementioned structural artifacts using an off-the-shelf DiFix3D model. However, because the base DiFix3D is pre-trained exclusively on static scene datasets, it suffers from a significant domain gap. Consequently, it fails to rectify the specific degradations induced by the multi-view inconsistencies of our generated canine data. By injecting a lightweight LoRA module trained on our self-supervised dataset and incorporating explicit geometric conditioning, \textbf{Model C} successfully adapts the generic 2D diffusion prior to these targeted degradation patterns, effectively eliminating geometric tearing and texture blurring.

\textbf{Effectiveness of the Progressive Self-Refinement Loop.} 
Although Model C successfully repairs structural artifacts in single-step renderings, applying this repair independently across multi-view images cannot guarantee absolute 3D spatial consistency, nor can it fully recover the high-frequency texture details of the input image. Our \textbf{Full Model} wraps this generative repair within an iterative closed-loop design. By systematically rendering, repairing, and re-optimizing the 3DGS parameters across three cycles, the self-refinement loop progressively distills authentic high-frequency details directly into the 3D geometry.

\section{Conclusion}
\label{sec:conclusion}

In this paper, we presented \textbf{CORGI}, a novel ``generation-then-reconstruction'' framework that tackles the highly challenging task of recovering high-fidelity 3D dogs from a single in-the-wild input image. To overcome the scarcity of multi-view canine data and the severe cross-view inconsistencies inherent from modern generative models, we strategically decomposed the problem into three components: governing canonical view synthesis, 3D lifting following dog prior, and geometry-guided texture refinement. Extensive evaluations demonstrate that CORGI achieves state-of-the-art reconstruction quality and generalizes seamlessly across a vast diversity of canine breeds and poses. By completely eliminating the need for paired 3D training data, our framework produces robust, artifact-free 3D models that are readily applicable to downstream animation and digital content creation.

Despite its promising results, CORGI has several limitations that point to fruitful directions for future research. First, our current animation pipeline relies on linear blend skinning (LBS) driven by the underlying D-SMAL template, which, while effective for canonical poses, struggles to produce fully realistic deformations for highly articulated or extreme poses. We believe this can be addressed by collecting and incorporating more diverse pose data to learn richer deformation priors. Second, our reconstruction framework is intrinsically tied to the D-SMAL prior, which assumes a canine-specific morphology. Extending CORGI to other quadruped species—or ideally, to arbitrary animal categories—remains an important avenue for future work, potentially by learning a more generalizable parametric animal prior from large-scale unlabeled imagery.

% \subsection*{Availability of data and materials}
% The data  are available from the corresponding author upon reasonable request.

% \subsection*{Acknowledgements}
% Acknowledgements of people,
% grants, funds, etc., should be placed in a separate section before reference
% list. The names of funding organizations should be written in full. Do not
% include acknowledgements on the title page, as a footnote to the title or
% otherwise.

% \subsection*{Declaration of competing interest}

% The authors have no competing interests to declare that are relevant to the
% content of this article.

% \subsection*{Electronic Supplementary Material  \note{(if applicable)}}

% If ESM is submitted, it will be published as received from the author in the
% online version only. ESM may consist of: (i) information that cannot be
% printed: animations, video clips, sound recordings; (ii) information that is
% more convenient in electronic form: sequences, spectral data, etc.; (iii)
% large amounts of original data, e.g., additional tables, illustrations, etc.
% If supplying any ESM, the text must make specific mention of the material as
% a citation, similar to that of figures and tables (e.g., Fig. S1 in the
% ESM). Besides, a paragraph should be added before the ``References'' section
% (e.g., Electronic Supplementary Material: Supplementary material (add a
% brief description) is available in the online version of this article).
% \pagebreak

% for bibtex
\bibliographystyle{CVMbib}
\bibliography{refs}

\subsection*{Author biography}
\vspace*{0.6em}

\begin{biography}[author1]{Yuxiao Wu} is a Ph.D. student in School of Mathematical Sciences, University of Science and Technology of China. His current research focuses on 3D animal reconstruction, single-image reconstruction and virtual reality.
\end{biography}
\vspace*{2.0em}

\begin{biography}[author2]{Weile Li} is an undergraduate student at the School of Artificial Intelligence and Data Science, University of Science and Technology of China. His current research interests include computer graphics, multimodal large models, and etc.
\end{biography}
\vspace*{2.0em}

\begin{biography}[author3]{Boyi Zhu} is an undergraduate student at the School of Mathematical Sciences, University of Science and Technology of China. His current research interests include computer graphics and 3D reconstruction.
\end{biography}
\vspace*{2.0em}

\begin{biography}[author4]{Yumeng Liu} is a Postdoctoral Researcher at the School of Mathematical Sciences, University of Science and Technology of China. She received her Ph.D. in Computer Science from the University of Hong Kong. Her research interests include 3D/4D reconstruction, computer graphics and embodied intelligence.
\end{biography}
\vspace*{2.0em}

\begin{biography}[author5]{Youcheng Cai} received the B.S. degree in information and computing science and Ph.D. degree in computer application technology from the Hefei University of Technology, Hefei, China, in 2008 and 2023, respectively. He is currently working as a postdoctor at University of Science and Technology of China. His research interests include 3D reconstruction, computer vision, and machine learning.
\end{biography}
\vspace*{2.0em}

\begin{biography}[author6]{XiaoMing Fu} received a BSc degree in 2011 and a Ph.D. in 2016 from the University of Science and Technology of China. He is an associate professor at the School of Mathematical Sciences, University of Science and Technology of China. His research interests include geometric processing and computer-aided geometric design. His research work can be found at his website: \url{http://staff.ustc.edu.cn/~fuxm/}geometric design.
\end{biography}
\vspace*{2.0em}

\begin{biography}[author7]{Ligang Liu}
received his B.Sc. and Ph.D. degrees from Zhejiang University, China, in 1996 and 2001, respectively. He is a professor at USTC. Between 2001 and 2004, he worked at Microsoft Research Asia, and he worked at Zhejiang University from 2004 to 2012. He paid academic visits to Harvard University in 2009 and 2011. His research interests include geometric processing and image processing.
\end{biography}
\vspace*{2.0em}

% \subsection*{Graphical abstract}

% It is a single, concise, and pictorial summary of the main findings of the article. It could either be the concluding figure from the article or better still a figure that is specially designed for the purpose, which captures the content of the article for readers at a single glance. We require it in JPG format, 300 dpi, and with the ratio of height to length $= 8 : 13$.

\end{document}